\documentclass[runningheads]{llncs}
\usepackage{graphicx}
\usepackage{amsmath}
\usepackage{amssymb}
\usepackage{amsfonts}
\usepackage{multirow}
\usepackage{arydshln}
\usepackage{color}
\usepackage{bm}

\makeatletter
\newcommand{\figcaption}[1]{\def\@captype{figure}\caption{#1}}
\newcommand{\tblcaption}[1]{\def\@captype{table}\caption{#1}}
\makeatother

\begin{document}

\title{Mitosis Detection from Partial Annotation \\ by Dataset Generation via Frame-Order Flipping}
\titlerunning{Mitosis Detection from Partial Annotation}

\author{Kazuya Nishimura\inst{1} \and
Ami Katanaya\inst{2} \and
Shinichiro Chuma\inst{2} \and
Ryoma Bise\inst{1}}


\authorrunning{K. Nishimura et al.}
%
\institute{Kyushu University, Fukuoka, Japan \email{kazuya.nishimura@humna.ait.kyushu-u.ac.jp} 
 \and Kyoto University, Kyoto, Japan}

\maketitle       
\begin{abstract}
Detection of mitosis events plays an important role in biomedical research.
Deep-learning-based mitosis detection methods have achieved outstanding performance with a certain amount of labeled data. However, these methods require annotations for each imaging condition. Collecting labeled data involves time-consuming human labor. 
In this paper, we propose a mitosis detection method that can be trained with partially annotated sequences.
The base idea is to generate a fully labeled dataset from the partial labels and train a mitosis detection model with the generated dataset.
First, we generate an image pair not containing mitosis events by frame-order flipping.
Then, we paste mitosis events to the image pair by alpha-blending pasting and generate a fully labeled dataset.
We demonstrate the performance of our method on four datasets, and we confirm that our method outperforms other comparisons which use partially labeled sequences.
Code is available at \url{https://github.com/naivete5656/MDPAFOF}.
\end{abstract}
\section{Introduction}
\begin{figure}[t]
 \centering
 \begin{tabular}{cc}
 \begin{minipage}{0.48\textwidth}
  \centering
  \includegraphics[width=0.9\linewidth]{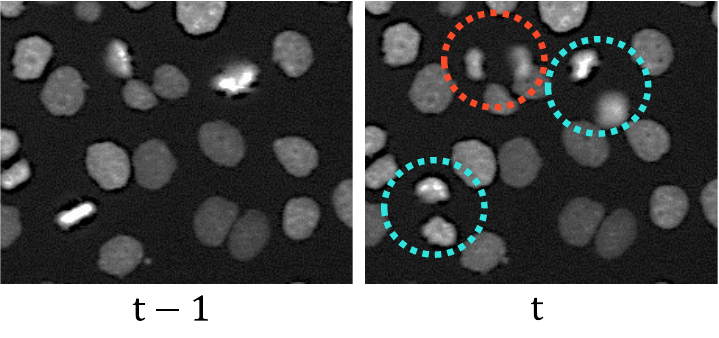}
  \caption{Example of partially labeled frames. The mitosis shown in the red-dotted circle is annotated, and the mitoses shown in the light-blue-dotted circles are not annotated.}
  \label{fig:intro}
 \end{minipage}%
 \begin{minipage}{0.04\textwidth}
  \hspace{0.04\columnwidth}
 \end{minipage}
 \begin{minipage}{0.48\textwidth}
  \centering
  \includegraphics[width=0.9\linewidth]{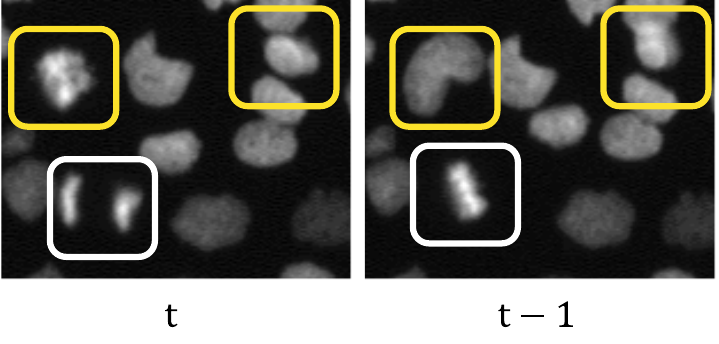}
  \caption{Example of frame-order flipped images. A mitosis event is converted into a merge event (white rectangles). Non-mitotic cells still have non-mitotic cell movement (yellow rectangles).}
  \label{fig:frame-flipped}
 \end{minipage}
 \end{tabular}
\end{figure}

Fluorescent microscopy is widely used to capture cell nuclei behavior. Mitosis detection is the task of detecting the moment of cell division from time-lapse images (the dotted circles in Fig. \ref{fig:intro}).
Mitosis detection from fluorescent sequences is important in biological research, medical diagnosis, and drug development. 

Conventionally tracking-based methods \cite{yang2005cell,thirusittampalam2013novel,debeir2005tracking,thirusittampalam2013novel} and tracking-free
methods \cite{gallardo2004mitotic,huh2010automated,gilad2019fully,liu2017multi} have been proposed for mitosis detection.
Recently, deep-learning-based mitosis-detection methods have achieved outstanding performance \cite{nishimura2020spatial,su2021spatio,nie20163d,mao2016hierarchical,mao2017two,lu2019sequential}. 
However, training deep-learning methods require a certain amount of annotation for each imaging condition, such as types of cells and microscopy and the density of cells. 
Collecting a sufficient number of labeled data covering the variability of cell type and cell density is time-consuming and labor-intensive. 

Unlike cell detection and segmentation, which aims to recognize objects from a single image, mitosis detection aims to identify events from time series of images. Thus, it is necessary to observe differences between multiple frames to make mitosis events annotation. Comprehensively annotating mitosis events is time-consuming, and annotators may be missed mitosis events. Thus, we must carefully review the annotations to ensure that they are comprehensive.

Partial annotation has been used as a way to reduce the annotation costs of cell and object detection \cite{fujii2021cell,yang2020object,qu2020weakly}. Fig. \ref{fig:intro} shows an example of partially annotated frames. Some mitosis events are annotated (a red-dotted circle), and others are not (light-blue-dotted circles). The annotation costs are low because the annotator only needs to plot a few mitotic positions. 
In addition, this style of annotation allows for missing annotations. Therefore, it would be effective for mitosis detection.

Unlike supervised annotation, partial annotation can not treat unannotated areas as regions not containing mitosis events since the regions may contain mitosis events (Fig. \ref{fig:intro}). The regions naturally affect the training in the partial annotation setting. To avoid the effect of unlabeled objects in unlabeled regions, Qu {\it et al.} \cite{qu2020weakly} proposed to use a Gaussian masked mean squared loss, which calculates the loss around the annotated regions. 
The loss function works in tasks in which foreground and background features have clearly different appearances, such as in cell detection. 
However, it does not work on mitosis detection since the appearance of several non-mitotic cells appears similar to mitosis cells; it produces many false positives.

In this paper, we propose a cell-mitosis detection method for fluorescent time-lapse images by generating a fully labeled dataset from partially annotated sequences. We achieve mitosis detection training in a mitosis detection model with the generated dataset. 
To generate the fully labeled dataset, we should consider two problems: (1) no label indicating regions not containing mitosis cells and (2) few mitosis annotations.

We can easily generate the regions not containing mitotic cells by using one image twice. 
However, such regions do not contribute to identifying mitotic cells and non-mitotic cells since the data do not show natural cell motions. 
For the training to be effective, the regions not containing mitotic cells should show the natural movements of cells. 
To generate such regions, we propose frame-order flipping which simply flips the frame order of a consecutive frame pair. As shown in the white rectangles in Fig. \ref{fig:frame-flipped}, we can convert a mitosis event to a cell fusion by flipping operation. Hence, the flipped pair is the region not containing mitosis cells. Even though we flipped the frame order, the non-mitotic cells still have natural time-series motion, as shown in the yellow rectangles in Fig. \ref{fig:frame-flipped}. 

In addition, we can make the most of a few partial annotations by using copy-and-paste-based techniques. Unlike regular copy-and-paste augmentation \cite{ghiasi2021simple} for supervised augmentation of instance segmentations which have object mask annotations, we only have point-level annotations. Thus, we propose to use alpha-blending pasting techniques which naturally blend two images. 

Experiments conducted on four types of fluorescent sequences demonstrate that the proposed method outperforms other methods which use partial labels.



\noindent
{\bf Related work}
Some methods used partially labeled data to train model \cite{qu2020weakly,yang2020object,fujii2021cell}.
Qu \cite{qu2020weakly} proposed a Gaussian masked mean squared loss, which calculates the loss around the annotated areas. To more accurately identify negative and positive samples, positive unlabeled learning has been used for object detection \cite{yang2020object,fujii2021cell}.
These methods have used positive unlabeled learning on candidates detected by using partial annotation to identify whether the candidates are labeled objects or backgrounds.
However, since the candidates detected by partial annotation include many false positives, the positive unlabeled learning does not work on mitosis detection.
the appearance of the mitosis event and backgrounds in the mitosis detection task, it is difficult to estimate positive prior. These methods could not work on mitosis detection.
The positive unlabeled learning requires a positive prior. 

\begin{figure}[t]
  \centering
   \includegraphics[width=0.9\linewidth]{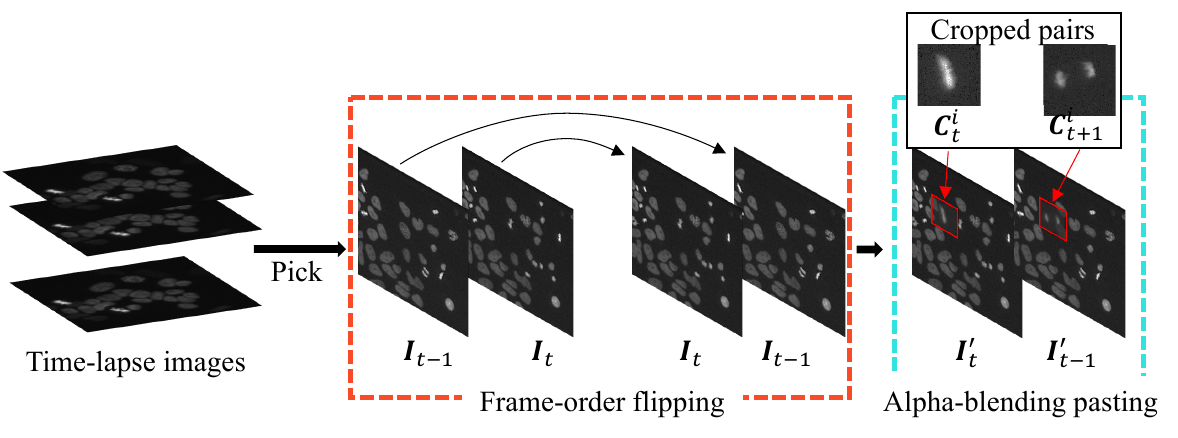}
    \caption{Overview of dataset generation.}
    \label{fig:overview}
\end{figure}

\section{Method: mitosis detection with partial labels}
Our method aims to detect coordinates and timing $(t, x, y)$ of mitosis events from fluorescent sequences.
For training, we use time-lapse images $\mathcal{I} = \{ \bm{I}_{t} \}_{t=1}^T$ and partial labels (a set of annotated mitosis cells). Here, $\bm{I}_{t}$ denotes an image at frame $t$, and $T$ is the total number of frames.

Our method generates a fully labeled dataset $\mathcal{D}_p= \{ (\bm{I}'_{t-1}, \bm{I}'_{t}), \mathcal{P}_t' \}^{T-1}_{t=1}$ from time-lapse images $\mathcal{I}$ and partial labels and then trains a mitosis detection model $f_{\theta}$ with the generated dataset. Here, $\bm{I}'_{t}$ is a generated image, and $\mathcal{P}_t'$ is a set of mitotic coordinates contained in $(\bm{I}'_{t-1}, \bm{I}'_{t})$. Since our method trains the network with partial labels, it can eliminate the costs of checking for missed annotations.

\subsection{Labeled dataset generation}
Fig. \ref{fig:overview} shows an overview of our dataset generation. We randomly pick a pair of consecutive frames $(\bm{I}_{t-1}, \bm{I}_t)$ from time-lapse images $\mathcal{I}$. Since the pair may contain unannotated mitosis events, we forcibly convert the pair into a negative pair ({\it i.e.,} a pair which does not contain mitosis events) by using frame-order flipping. Next, we paste mitosis events to a generated pair using alpha-blending pasting and obtain a generated pair $(\bm{I}'_{t-1}, \bm{I}'_{t})$. Since we know the pasted location, we can obtain the mitosis locations $\mathcal{P}'_t$ of the generated pair.

\begin{figure}[t]
    \centering
    \includegraphics[width=0.85\linewidth]{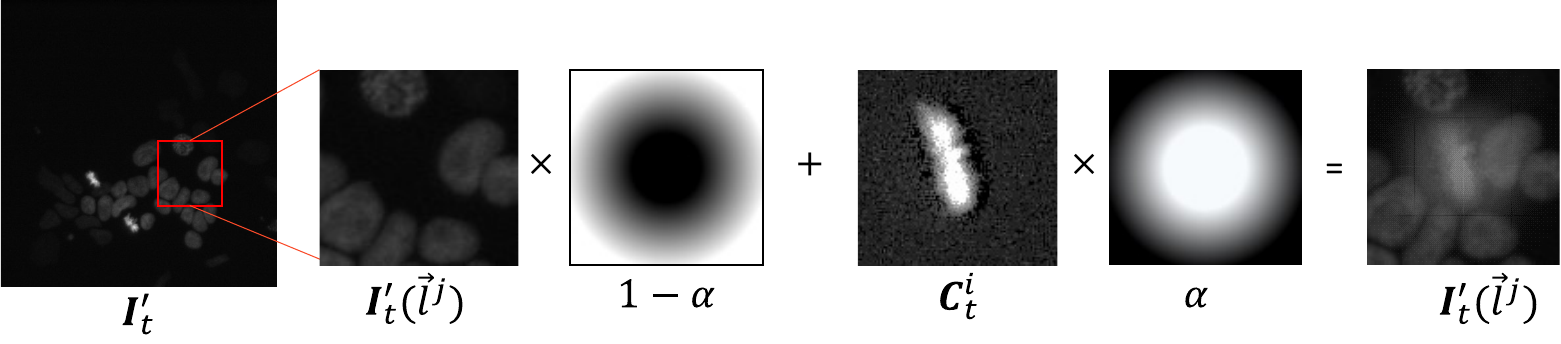}
    \caption{Illustration of alpha-blending pasting.}
    \label{fig:paste}
\end{figure}

\noindent
{\bf Negative pair generation with frame-order flipping:} 
In this step, we generate a pair not containing mitotic cells by using a simple augmentation-based frame-order flipping. Fig. \ref{fig:overview} shows an example of the pair images $(\bm{I}_{t-1}, \bm{I}_t)$. The pair may contain mitosis events. If we assume that the pair does not contain mitotic cells, it affects the training of the mitosis detection model $f_{\theta}$. To prevent the pair from containing mitosis events, we flip the frame order and treat the flipped pair $(\bm{I}_{t}, \bm{I}_{t-1})$ as a pair of negative.

Since mitosis is the event that a cell divides into two daughter cells, the mitosis event is transformed into an event in which two cells fuse into one by flipping the order (Fig. \ref{fig:frame-flipped}).
The flipped event can treat as a non-mitotic event.
Note that the motivation behind using frame flipping is to be able to utilize pixels showing the motions of non-mitotic cells negatives by transforming mitosis into other events.
Even if the order is flipped, the movements of the non-mitotic cell are still a non-mitotic cell feature, and we consider that these cells are effective for the training of the negative label.

\noindent
{\bf Mitosis label utilization with alpha-blending pasting:} 
Next, we paste mitosis events to the flipped pair by using copy-and-paste techniques in order to utilize the positive labels effectively. 
Copy and paste augmentation has been used for supervised augmentation of instance segmentation \cite{ghiasi2021simple}.
Unlike instance segmentation with object masks, we only have locations (t, x, y). 
A simple solution is to crop images around the mitosis position and copy and paste them to the target image, like in CutMix \cite{yun2019cutmix}. However, the cropped image naturally contains surrounding objects, and the generated image appears unnatural. Unnatural images cause the detection network to make biased predictions and reduce generalization performance. 
To avoid this problem, we propose alpha-blending pasting with a Gaussian blending mask.
We blend two images by leaving the pixel value in the center and blurring the vicinity of the edge of the image.

First, we crop the image around the positive annotations and obtain a set of cropped pair $\mathcal{C} = \{(\bm{C}_{t-1}^i, \bm{C}_{t}^i )\}^N_{i=0}$ and initialize $(\bm{I}'_{t-1}, \bm{I}'_{t})=(\bm{I}_{t}, \bm{I}_{t-1})$ and $\mathcal{P}_t'= \{ \}$. Here, $N$ is the total number of partial annotations, while $\bm{C}_{t-1}^i$ and $\bm{C}_{t}^i$ are images before and after the mitosis of the $i$-th annotation (Fig. \ref{fig:overview}). Define $\bm{I}_{t}'(\vec{l}^j)$, $\bm{I}_{t-1}'(\vec{l}^j)$ as a cropped pair image at the $j$-th random spatial location $\vec{l}^j$. 
We crop each image centered at $\vec{l}^j$ to a size that is the same as that of $\bm{C}_{t}^i$. We update the randomly selected patch $\bm{I}_{t}'(\vec{l}^j)$, $\bm{I}_{t-1}'(\vec{l}^j)$ by blending a randomly selected cropped pair $(\bm{C}_{t-1}^i, \bm{C}_{t}^i)$ with the following formula: $\bm{I}_{t}'(\vec{l}^j) = (1-\alpha) \odot \bm{I}_{t}'(\vec{l}^j) + \alpha \odot \bm{C}_{t}^i$, where $\alpha$ is a Gaussian blending mask (Fig. \ref{fig:paste}). 
We generate the blending mask by blurring a binary mask around the annotation with a Gaussian filter. We use a random sigma value for the Gaussian filter. Then, we add the paste location $\vec{l}^j$ to the set $\mathcal{P}_t'$. We repeat this process random $k$ times. 

\subsection{Mitosis detection with generated dataset}
\begin{figure}[t]
  \centering
  \begin{tabular}{cc}
  \begin{minipage}{0.48\hsize}
    \centering
    \includegraphics[width=0.93\linewidth]{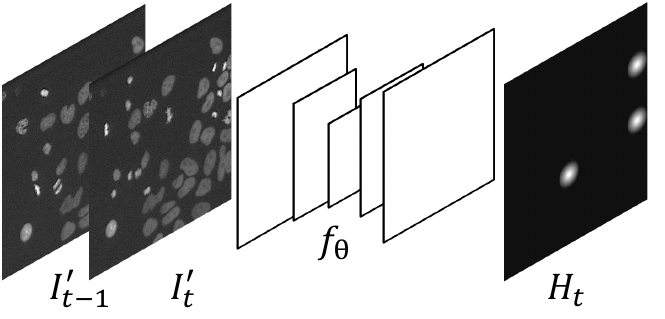}
    \caption{Cell mitosis detection model.}
    \label{fig:baseline}
  \end{minipage}%
  \begin{minipage}{0.48\textwidth}
    \centering
    \includegraphics[width=0.93\linewidth]{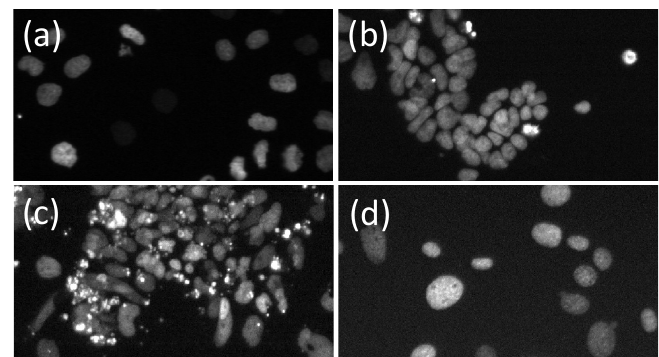}
    \caption{Example of images in datasets. (a) HeLa, (b) ES, (c) ES-D, (d) Fib.}
    \label{fig:dataset}
  \end{minipage}
  \end{tabular}
\end{figure}

We modified a heatmap-based cell detection method \cite{nishimura2021weakly} to work as a mitosis detection method. 
Fig. \ref{fig:baseline} is an illustration of our mitosis detection model. 
Given two consecutive frames $(\bm{I}'_{t-1}, \bm{I}'_{t})$, the network output heatmap $\hat{\bm{H}}_{t}$. 
We treat the channel axis as the time axis for the input.
The first channel is $\bm{I}'_{t-1}$, and the second is $\bm{I}'_{t}$.

First, we generate individual heatmaps $\bm{H}_{t}^j$ for each pasted coordinate $\vec{l}^j = (l^j_x, l^j_y)$. $\bm{H}_{t}^j$ is defined as $\bm{H}_{t}^j(p_x, p_y) = \exp \left( -\frac{(l_x^j - p_x) ^2 + (l_y^j - p_y) ^ 2}{\sigma^2} \right)$, where $p_x$ and $p_y$ are the coordinates of $\bm{H}_{t}^j$ and $\sigma$ is a hyper parameter that controls the spread of the peak. 
The ground truth of the heatmap at $t$ is generated by taking the maximum through the individual heatmaps, $\bm{H}_{t} = \max_j (\bm{H}^j_t)$ ($\bm{H}_{t}$ in Fig. \ref{fig:baseline}). The network is trained with the mean square error loss between the ground truth $\bm{H}_t$ and the output of the network $\hat{\bm{H}}_t$. We can find the mitosis position by finding a local maximum of the heatmap.

\section{Experiments}
\noindent
{\bf Dataset:} 
We evaluated our method on four datasets. 
The first set is {\bf HeLa} \cite{ulman2017objective}, in which live cell images of HeLa cells expressing H2B-GFP were captured with 1100 $\times$ 700 resolution \cite{ulman2017objective} \footnote{We used the publicly available
CTC data-set \url{http://celltrackingchallenge.net/}. We only use HeLa since the number of mitosis events in other cells is small.}. 
Each sequence contains 92 fluorescent images with 141 mitosis events on average.
The second set is {\bf ES}, in which live cell images of mouse embryonic stem cells expressing H2B-mCherry were captured with 1024 $\times$ 1024 resolution.
Each sequence contains 41 fluorescent images with 33 mitosis events on average.
The third set is {\bf ES-D} in which mouse embryonic stem cells expressing H2B-mCherry were induced to differentiate and used to capture live cell images.
Each sequence contains 61 fluorescent images with 18 on average events on average.
The fourth set is {\bf Fib}, in which live cell images of mouse fibroblast cells expressing H2B-mCherry were captured with 1024 $\times$ 1024 resolution.
Each sequence contains 42 fluorescent images with 11 mitosis events on average.
Each dataset consists of four sequences of images.
We performed four-fold cross-validation in which two sequences were used as training data, one as validation data, and one as test data.
As shown in Fig. \ref{fig:dataset}, the appearance and density are different depending on the dataset.

\noindent
{\bf Implementation details:} 
We implemented our method within the Pytorch framework \cite{paszke2019pytorch} and used a UNet-based architecture \cite{ronneberger2015u} for the mitosis-detection network. The model was trained with the Adam optimizer with a learning rate of 1e-3. $\sigma$, which controls the spread of the heatmap, was 6. 
The cropping size of the positive annotations was $40$ pixels. 
We randomly change the number of pasting operations $k$ between $1$ and $10$. 
We used random flipping, random cropping, and brightness change for the augmentation.

\begin{table}[t]
  \caption{Quantitative evaluation results (F1-score).}
  \centering
    \label{tab:quan}
    \begin{tabular}{c| ccccc |ccccc} \hline
        &    \multicolumn{5}{c|}{1-shot} & \multicolumn{5}{c}{5-shot} \\
        Method  & HeLa & ES & ES-D & Fib & Ave. &HeLa & ES & ES-D & Fib & Ave. \\ \hline \hline
        Baseline \cite{nishimura2021weakly} & 0.356&0.581&0.06&0.235& 0.308 & 0.591 & 0.756 & 0.277 & 0.210 & 0.459 \\ 
        GM \cite{qu2020weakly} & 0.315&0.303&0.057&0.119& 0.199& 0.501 & 0.523 & 0.123 & 0.230 & 0.344 \\ 
        PU \cite{yang2020object} & 0.030 & 0.095 & 0.012 & 0.053 & 0.048& 0.463 & 0.538 & 0.375& 0.224 & 0.400 \\ 
        PU-I \cite{fujii2021cell} & 0.499 & 0.177 & 0.035 & 0.115& 0.207& 0.474 & 0.420 & 0.037 & 0.141 & 0.268 \\ 
        Ours & {\bf 0.593} & {\bf 0.740} & {\bf 0.439} & {\bf 0.440}& {\bf 0.553}&  {\bf 0.795} & {\bf 0.843} & {\bf 0.628} & {\bf 0.451} & {\bf 0.610}\\ \hline
    \end{tabular}
\end{table}

\noindent
{\bf Evaluation metrics:}
We evaluated our method using the F1 score \cite{su2021spatio}, which is widely used in mitosis detection. Given ground-truth coordinates and detected coordinates, we performed one-by-one matching. If the distance of the matched pair was within spatially 15 pixels and temporally 6, we associated the closest coordinate pairs. We treated the matched pair as true positives (TP), unassociated coordinates as false positives (FP), and unassociated ground-truth coordinates as false negatives (FN).

\begin{figure}[t]
    \centering
    \begin{tabular}{cc}
      \begin{minipage}{0.4\linewidth}
      \centering
      \tblcaption{Ablation study. FOF: frame-order flipping, ABP: alpha-blending pasting.}
        \label{tab:ablation}
        \begin{tabular}{c|c} \hline
            Method & F1 \\ \hline \hline
            w/o FOF &0.570 \\ 
            w/o ABP \cite{yun2019cutmix} &0.670 \\  
            Ours &\bf{0.795} \\ \hline 
        \end{tabular}
  \end{minipage}%

  \begin{minipage}{0.6\linewidth}
  \centering
    \includegraphics[width=0.8\linewidth]{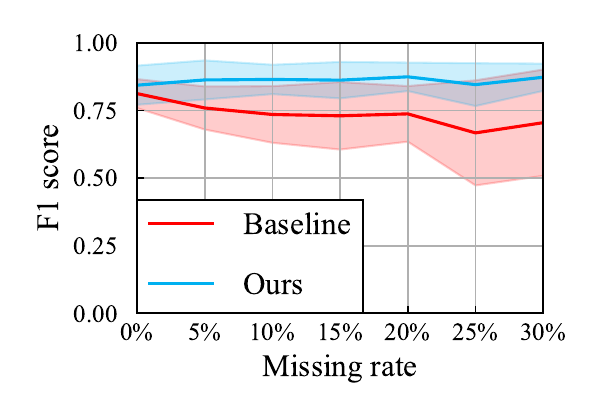}
    \caption{Robustness against missing annotations.}
    \label{fig:missing}
  \end{minipage}
  \end{tabular}
\end{figure}

\noindent
{\bf Comparisons:}
We conducted four comparisons that involved training the model with partially labeled data. For the first method, we trained the model by treating unlabeled pixels as non-mitosis ones (Baseline \cite{nishimura2021weakly}). The second method used the Gaussian masked loss (GM \cite{qu2020weakly}). The masked loss was calculated on the masked pixels around the positive-label pixels. 
Thus, the method ignored unlabeled pixels. The third method used positive unlabeled learning to identify mitosis from candidates obtained by the detection model trained with the masked loss (PU \cite{yang2020object}). 
The fourth method generated pseudo-labels from the results of positive unlabeled learning and retrained the detection model with the pseudo-label (PU-I \cite{fujii2021cell}). 

In Table \ref{tab:quan}, we compared our method with previous methods in one and five-shot settings. 
We used N samples per sequence in the N-shot settings. For a robust comparison, we sampled one or five mitosis annotations under five seed conditions and took the average. 
Overall, our method outperformed all compared methods in F1 metric. GM \cite{qu2020weakly}, PU \cite{yang2020object}, and PU-I \cite{fujii2021cell} are designed for detecting objects against simple backgrounds. Therefore, these methods are not suited to a mitosis detection task and are inferior to the baseline.

The baseline \cite{nishimura2021weakly} treats unlabeled pixels as non-mitosis cell pixels. 
In the partially labeled setting, unlabeled pixels contain unannotated mitosis events, and unannotated mitosis affects performance. 
Unlike cell detection, mitosis detection requires identifying mitosis events from various non-mitotic cell motions, including motions that appear mitotic appearances. Although GM \cite{qu2020weakly} can ignore unlabeled mitosis pixels with the masked loss, it is difficult to identify such non-mitosis motions. 
Therefore, GM estimates produce many false positives. PU \cite{yang2020object} uses positive unlabeled learning to eliminate false positives from candidates obtained from the detection results with partial labels. However, positive unlabeled learning requires a positive prior in the candidates and a certain amount of randomly sampled positive samples. Since the candidates contain many false positives, the positive prior is difficult to estimate. In addition, there is no guarantee that positive unlabeled learning can work correctly with the selected N-shot annotations. 
Moreover, since positive unlabeled learning does not work in the mitosis detection task, PU-I \cite{fujii2021cell} can not select accurate pseudo labels.
Unlike these methods, our method can estimate mitosis events accurately. Since our method generates a fully labeled dataset from a partial label, it effectively uses a few partial annotations.

\begin{figure}[t]
    \begin{tabular}{cc}
  \begin{minipage}{0.5\linewidth}
    \centering
    \includegraphics[width=0.9\linewidth]{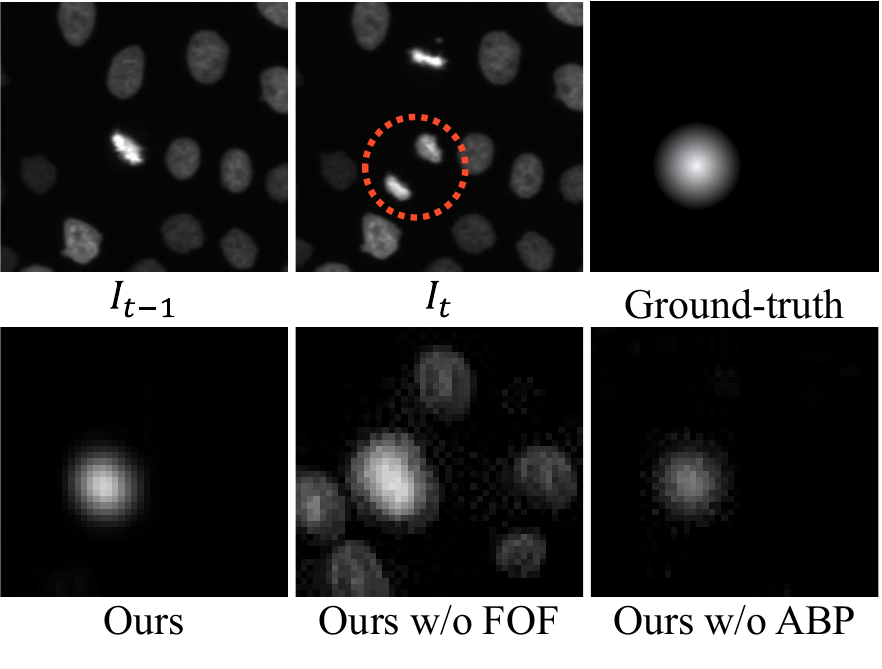}
    \caption{Example of estimation results. Red dotted circles indicate mitosis events.}
    \label{fig:exampleresult}
  \end{minipage}
  \begin{minipage}{0.5\linewidth}
    \centering
    \includegraphics[width=0.9\linewidth]{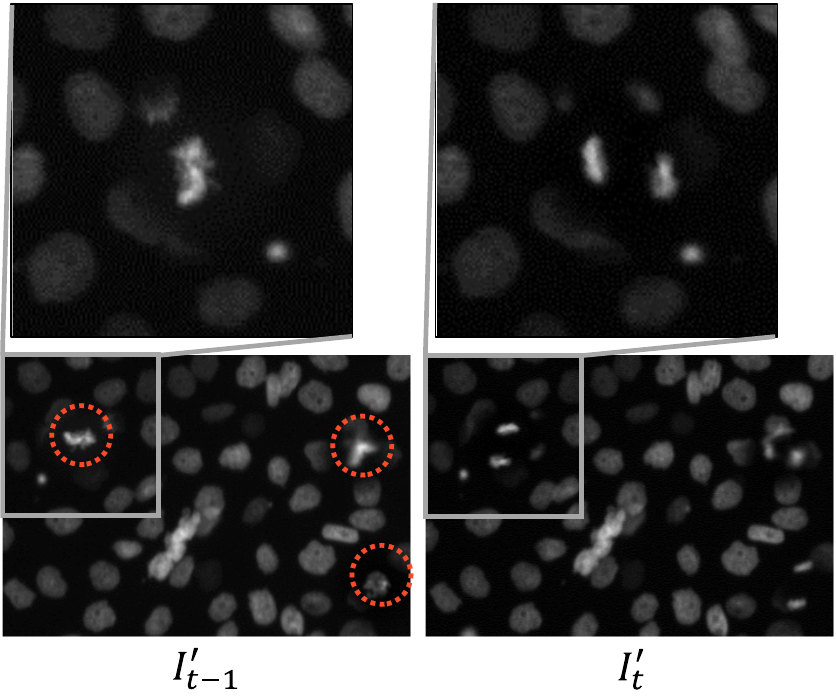}
    \caption{Example of generated images.}
    \label{fig:generated}
  \end{minipage}%
  \end{tabular}
\end{figure}

\noindent
{\bf Effectiveness of each module:}
We performed an ablation study on the HeLa dataset to investigate the effectiveness of the proposed module.
We used random augmentation ({\it i.e.,} random elastic transformation \cite{simard2003best}, brightness change, and gaussian noise) instead of using frame-order flipping (FOF).
We generated $\bm{I}_t^{aug}$ by augmenting $\bm{I}_t$ and input the pair $(\bm{I}_t, \bm{I}_t^{aug})$ to the network. 
In the w/o ABP setting, we directly pasted cropped images on the target image as in CutMix \cite{yun2019cutmix}.
Table \ref{tab:ablation} demonstrates that the proposed modules improve mitosis detection performance.
Fig. \ref{fig:exampleresult} shows examples of the estimation results for each condition. 
Without the FOF setting, the detection model estimates a high value for all moving cells, leading to over-detection.
Without the ABP setting, the detection model overfits the directly pasted image. 
The directly pasted image tends to include unnatural boundaries on the edge, leading to missed detections in real images.

\noindent
{\bf Robustness against missing annotations:}
We confirmed the robustness of the proposed method against missing annotations on the ES dataset. We changed the missing annotation rate from $0\%$ to $30\%$.
A comparison with the supervised method in terms of F1-score is shown in Fig. \ref{fig:missing}.
The performance of the supervised method deteriorates as the percentage of missing labels increases, whereas the performance of the proposed method remains steady. Since our method flips the frame order, we can avoid the effects of missing annotations.

\noindent
{\bf Appearance of generated dataset:}
Fig. \ref{fig:generated} shows an example of the generated image pair. The cropped mitosis image pairs were pasted on the red-dotted circle.
It can be seen that the borders of the original image and the pasted image have been synthesized very naturally.

\section{Conclusion}
We proposed a mitosis detection method using partially labeled sequences with frame-order flipping and alpha-blending pasting. Our frame-order flipping transforms unlabeled data into non-mitosis labeled data through a simple flipping operation. Moreover, we generate various positive labels with a few positive labels by using alpha-blending pasting. Unlike directly using copy-and-paste, our method generates a natural image. Experiments demonstrated that our method outperforms other methods that use partially annotated sequences on four fluorescent microscopy images.

\noindent
{\bf Acknowledgements:} This work was supported by JSPS KAKENHI Grant Number JP21J21810 and JST ACT-X Grant Number JPMJAX21AK, Japan. 
%
%

\bibliographystyle{splncs04}
\bibliography{mybibliography}

\end{document}